\documentclass[pdflatex,sn-mathphys-num]{sn-jnl}
\usepackage{graphicx}%
\usepackage{multirow}%
\usepackage{amsmath,amssymb,amsfonts}%
\usepackage{amsthm}%
\usepackage{mathrsfs}%
\usepackage[title]{appendix}%
\usepackage{xcolor}%
\usepackage{textcomp}%
\usepackage{algorithm}%
\usepackage{algorithmicx}%
\usepackage{algpseudocode}%
\usepackage{booktabs}%
\usepackage{listings}%
\usepackage{array}
\usepackage{wrapfig} 
\usepackage[font=small]{caption} 
\usepackage[font=small]{subcaption}
\usepackage{fancyhdr}
\begin{document}

\title[Article Title]{Labeling Free-text Data using Language Model Ensembles}


\author*[1]{\fnm{Jiaxing}\sur{Qiu} }\email{jq2uw@virginia.edu}

\author[1]{\fnm{Dongliang}\sur{Guo} }

\author[2]{\fnm{Natalie}\sur{Papini}}

\author[3]{\fnm{Noelle}\sur{Peace}}

\author[3]{\fnm{Hannah F.}\sur{Fitterman-Harris}}

\author[3]{\fnm{Cheri A.}\sur{Levinson}}

\author[1]{\fnm{Tom}\sur{Hartvigsen} }

\author[1,4]{\fnm{Teague R.}\sur{Henry} }

\affil[1]{\orgname{University of Virginia, School of Data Science}}

\affil[2]{\orgname{Northern Arizona University, College of Health and Human Services}}

\affil[3]{\orgname{University of Louisville, Department of Psychology and Brain Sciences}}

\affil[4]{\orgname{University of Virginia, Department of Psychology}}

\abstract{ 

Free-text responses are commonly collected in psychological studies, providing rich qualitative insights that quantitative measures may not capture. Labeling curated topics of research interest in free-text data by multiple trained human raters is typically labor-intensive and time-consuming. Though large language models (LLMs) excel in language-based tasks, LLM-assisted labeling techniques relying on closed-source LLMs cannot be directly applied to clinical free-text data, without explicit consent for external use.

Analogous to labeling by multiple human raters, we propose a framework of assembling locally-deployable LLMs to improve the process of the labeling of predetermined topics in free-text data under privacy constraints. This framework leverages the heterogeneity of diverse open-source LLMs, and the ensemble approach seeks a balance between the agreement and disagreement across LLMs, guided by a relevancy scoring methodology that utilizes embedding distances between topic descriptions and LLMs' reasoning. We evaluated the ensemble approach using both publicly accessible Reddit data from eating disorder related forums, and free-text responses from eating disorder patients, both complemented by human annotations.

We found that: (1) there is heterogeneity in the performance of labeling among same-sized LLMs, with some showing low sensitivity but high precision, while others exhibit high sensitivity but low precision. (2) Compared to individual LLMs, the ensemble of LLMs achieved the highest accuracy and optimal precision-sensitivity trade-off, as evaluated by human annotations. (3) The relevancy scores across LLMs showed greater agreement than dichotomous labels, indicating that the relevancy scoring method effectively mitigates the heterogeneity in LLMs' labeling.

}


\maketitle

\newpage

\section{Introduction}\label{sec_intro}

In psychology, free-form text responses and qualitative data have been effectively collected since the beginning of scientific inquiry \cite{sarkhel_kaplan_2009, hua_large_2024}. To perform quantitative analyses on psychological textual data, curated topics such as symptom manifestations \cite{cook_novel_2016}, psychosis risk factors \cite{irving_using_2021}, social risk factors \cite{skaik_using_2020}, and pathological behaviors \cite{levis_natural_2021} must first be labeled (coded) and tailored to specific research interests. Conventionally, this process involves multiple human coders independently labeling topics of interest, with reliability evaluated through the level of agreement \cite{tinsley_interrater_1975}. However, this process is typically time-consuming and labor-intensive, involving both the careful training of human coders as well as their efforts in reading and labeling large volumes of text.

Pre-trained large language models (LLMs) are statistical models that can be adapted for a variety of downstream tasks. Research have shown that larger-sized LLMs, such as the GPTs \cite{brown_language_2020, openai_gpt-4_2024, shahriar2024putting}, offer significant opportunities in data annotation tasks \cite{tan_large_2024, tai_examination_2024, gilardi_chatgpt_2023, alizadeh_open-source_2024, chew_llm-assisted_2023}. However, ethical concerns regarding privacy and safety of free-text data should be central to the development of LLM-assisted methodologies \cite{hua_large_2024, cabrera_ethical_2023, kacetl_legislative_2016}. Without explicit consent for external use, closed-source LLMs, or large-sized LLMs that cannot be deployed locally, are often unsuitable for direct application to free-text data collected in clinical or psychological research settings \cite{cabrera_ethical_2023, kacetl_legislative_2016, kjell_beyond_2024}.

In this study, we focus on enhancing the accuracy and reliability of labeling curated topics on free-text data, while accounting for both security and computational constraints in clinical and psychological research. Instead of inferring psychological conditions or constructs, our work focus on labeling specific topics that require detailed descriptions tailored to individual research interests.
Analogous to the labeling process by multiple human raters, we propose a framework that assembles multiple small-sized (7–8 billion parameter), locally deployable language models to label and score the presence of target topics in free-text data.

The rest of the article is organized as follows: We dedicate the remainder of the introduction to detailing the background of LLMs (\ref{sec_intro_part1}) and their applications in data annotation generally, and specifically in psychology (Section \ref{sec_intro_part2}). We then discuss the common ensemble technique and related work to motivate the use of locally deployable LLM ensembles for labeling target topics in free-text data (Section \ref{sec_intro_part3}). In the methods section (Section \ref{sec_method}), we introduce our ensemble framework (Section \ref{sec_method_ensemble}), and detail the evaluations using both a public and private psychological dataset (Section \ref{sec_method_evaluation}). Section \ref{sec_result} compares the ensemble with individual LLMs in terms of fidelity to human labeling. Section \ref{sec_discussion} contains a discussion of our proposed framework.

\subsection{Background of Large Language Models}\label{sec_intro_part1}

Large language models are large-scale, pre-trained neural networks known for their capacity and versatility in processing and generating human-like language \cite{liu_pre-train_2021}. 
Popular LLMs typically rely on the Transformer architecture and its attention mechanism \cite{vaswani_attention_2023}, which helps to encode not only the semantic meaning of words but also their relationships and surrounding context. 
Natural language inputs are first transformed into continuous vectors that encode semantic and positional information. These embeddings are then typically processed by an encoder–decoder (sequence‑to‑sequence) \cite{devlin2019bert, vaswani_attention_2023, warner2024smarter} or a decoder‑only (autoregressive) architecture \cite{shaw2018self, radford2018improving, brown2020language} to generate the most probable next token. 
LLMs can be adapted for a variety of downstream tasks including text classification \cite{howard2018universal, sarzynska2021detecting, wang2018glue, wang2019superglue}, topic modeling \cite{angelov2020top2vec, grootendorst2022bertopic}, and question answering \cite{shuster2021retrieval, izacard2023atlas, bianchi2021crosslingualcontextualizedtopicmodels, lewis2020retrieval}. 
For language-based usages, LLMs can be broadly categorized into models designed for representation learning and models designed for generation. Representation-focused models, such as encoder-only or encoder–decoder architectures (e.g., BERT \cite{devlin2019bert}, RoBERTa \cite{liu2019roberta}), are primarily trained to produce contextual embeddings for downstream tasks like classification or retrieval. In contrast, generative models, typically decoder-only architectures (e.g., GPTs \cite{brown_language_2020, openai_gpt-4_2024, shahriar2024putting}), are trained to autoregressively generate coherent and human-like language responses.

Importantly, with respect to accessibility and security, LLMs can be categorized as open-source or closed-source depending on developers’ policies \cite{hua_large_2024}. Open-source models, such as LLaMA \cite{touvron_llama_2023}, support local deployment, enabling dataset customization and ensuring data privacy \cite{liu_pre-train_2021}. In contrast, closed-source models, such as commercial GPTs \cite{brown_language_2020, openai_gpt-4_2024, shahriar2024putting}, are controlled by third parties and accessed through application programming interfaces (APIs).


\subsection{Language Models for Data Annotation}\label{sec_intro_part2}
Research has shown that advanced LLMs present promising opportunities for facilitating data annotation \cite{tan_large_2024, tai_examination_2024, gilardi_chatgpt_2023, alizadeh_open-source_2024, chew_llm-assisted_2023}. The use of larger-sized and closed-source LLMs typically outperform smaller-sized and open-source LLMs. \citet{gilardi_chatgpt_2023} demonstrated that ChatGPT outperforms open-source LLMs in generic annotation tasks, including relevance, stance, topic, and frame detection. \citet{alizadeh_open-source_2024} demonstrated that fine-tuning can enhance the performance of open-source LLMs in text classification tasks within political science, getting closer to the performance of zero-shot prompting with GPT-3.5 and GPT-4. \citet{chew_llm-assisted_2023} proposed an LLM-assisted content analysis approach using GPT-3.5, which involves iterative steps by both humans and the LLM: codebook development, labeling, and agreement evaluation, repeated until a satisfactory level of human-LLM agreement is achieved. They found that GPT-3.5 can often perform labeling at agreement levels comparable to those of human coders. 

In psychology, LLMs have also been employed to annotate psychological content, along with tasks such as data augmentation \cite{ye_llm-da_2024, ding_data_2024}, resource enrichment \cite{qiu_benchmark_2023}, and the development of new LLM agents \cite{xu_mental-llm_2024, lai_psy-llm_2023}.
Specifically, LLMs have been used as classification models to label the presence or absence of general mental health conditions (e.g., suicide and depression \cite{lamichhane_evaluation_2023, yang_mentallama_2024}), as well as to provide more detailed information about these conditions (e.g., severity levels and subtypes \cite{yang_towards_2023, zhou_identifying_2023, hua_large_2024}). Notably, \textbf{closed-source LLMs} such as GPTs are at the forefront of these applications \cite{hua_large_2024, zhang_symptom_2022}.
Additionally, closed-source LLMs have been employed to annotate psychological constructs. \citet{rathje_gpt_nodate} demonstrated that GPT-3.5 and GPT-4 can accurately identify constructs such as sentiment, discrete emotions, and moral foundations across multiple languages using simple prompts without additional training, with performance validated by human annotators. Similarly, \citet{peters_large_2024} showed that GPT-3.5 and GPT-4 can infer personality traits and psychological dispositions from social media data. Extending further, \citet{kjell_beyond_2024} provided evidence that, with careful validation of deployment scenarios, the latest LLM technologies can shift psychological assessment from traditional rating scales toward natural language communication.

However, several considerations must be addressed before using LLMs to label curated psychological topics in free-text data collected from clinical or research settings. 

First, as a comprehensive review by \citet{hua_large_2024} pointed out, labels of mental health conditions in existing datasets are typically not curated, with substantial variability in their definitions across studies. Curated psychological topics tailored to individual research needs—such as pathological behaviors, social or cultural factors related to a mental condition in a certain population, and potential risk factors hypothesized by researchers—should be differentiated from clinical diagnoses of mental health conditions (e.g., depression or suicide) and the assessment of abstract constructs (e.g., sentiment or personality traits). Given the diversity of such curated topics and their study-specific definitions, evaluating LLMs' performance demands a higher level of precision, complemented with human annotations tailored to each topic's definition. In this study, we focused on curated topics with detailed definitions, ensuring that human annotations in our empirical evaluations closely aligned with the LLMs' labeling process.

Second, ethical concerns regarding the privacy and safety of free-text data are pivotal in developing LLM-assisted methodologies \cite{hua_large_2024, cabrera_ethical_2023, kacetl_legislative_2016}. Current applications frequently utilize closed-source, large-scale LLMs, such as GPTs. While these models provide strong language processing capabilities, they require data to be sent beyond the researchers' control. Sharing protected health-related data (such as clinician notes) via APIs and storing it in non-private environments, whether temporarily or long-term, poses risks to data privacy and safety. Moreover, without explicit consent for external use—which is rarely obtained if at all—closed-source LLMs are unsuitable for direct application to free-text data collected in clinical or research settings \cite{cabrera_ethical_2023, kacetl_legislative_2016, kjell_beyond_2024}. Larger-sized LLMs that can hardly be deployed locally in a secure computing environment, such as the open-source Llama-3.1-405B \cite{patterson_carbon_2022}, may also be unsuitable for handling protected free-text data. Thus, our work focuses on improving the accuracy and reliability of topic labeling by assembling diverse, open-source, small-sized (7-8 billion parameters), and locally-deployable LLMs, under security and computational constraints.

Lastly, the use of a single LLM in data annotation tasks is commonplace in current applications. However, studies have shown that diverse LLMs exhibit varying capacities across tasks \cite{ziems_can_2024, xu_llms_2024}, and we should expect heterogeneity in their performance when annotating diverse topics with granular definitions. As highlighted in a survey on LLM-assisted data annotation \cite{tan_large_2024}, employing the Mixture of Experts concept \cite{jordan1994hierarchical} and ensembling multiple LLMs can leverage this heterogeneity to enhance overall performance and computational efficiency \cite{artetxe_efficient_2022}, which we elaborate on in the next section.

\subsection{LLM Ensembles for Labeling Curated Topics in Free-text Data} \label{sec_intro_part3}


Studies have shown that diverse LLMs exhibit varying capacities in performing inference tasks such as coding, taxonomic labeling, and math \cite{ziems_can_2024, xu_llms_2024, satpute2024can}. We expect to observe heterogeneity in the performances of diverse open-source, same-sized LLMs when labeling various curated psychological topics. Besides differences in parameter sizes and model architectures \cite{kaplan_scaling_2020}, the heterogeneity in performance among LLMs also stems from the varied training datasets and preprocessing steps each LLM was exposed to, or inherited if the LLM is a variant of another \cite{raffel_exploring_2020}.

The main objective of ensembling diverse LLMs is to leverage the heterogeneity in LLMs and achieve reliable performance in topic labeling tasks. Ensemble learning of LLMs is typically performed either by ensembling model weights or by combining diverse outputs \cite{jacobs_adaptive_1991, jiang_llm-blender_2023}. A common technique–LLM routing–involves training a router model using various benchmark datasets to assign LLMs to their specialized tasks and applications, aiming to balance performance and cost \cite{lu_routing_2023, jiang_llm-blender_2023, chen_frugalgpt_2023, hu_routerbench_2024, jiang_mixtral_2024, ong_routellm_2024}. However, this technique is not directly applicable to our task of curated topic labeling for free-text data. (1) Current router models require the inclusion of large-sized, closed-source LLMs such as GPT at the high-cost, high-performance end of the performance-cost trade-off \cite{jiang_llm-blender_2023, chen_frugalgpt_2023, jiang_mixtral_2024}, which is not directly applicable to free-text data. (2) The development of router models has primarily focused on assigning LLMs to their specialties in generic tasks, such as commonsense reasoning, conversation, mathematics, and code generation. Routing LLMs for the specific task of topic labeling has not been investigated, except for one study that will be discussed in detail in the related work section. The effectiveness of using this generic technique to label curated psychological topics remains unestablished. (3) Training a router model requires a pre-specified collection of LLMs and extensive training datasets, making it challenging to scale to new LLM selections, as well as diverse datasets and research needs in psychology studies.

Therefore, in this study, we propose an ensembling framework tailored to the task of curated topic labeling in free-text data. \textbf{Our use of multiple LLMs is analogous to employing multiple human raters}, who are trained to label curated topics for specific research questions but are nevertheless constrained by their individual knowledge and biases. Importantly, we focus on the task of identifying predetermined topics of interest (a top-down approach), which is common in human annotation, rather than a data-driven exploratory topic analysis task using methods such as Latent Dirichlet Allocation (a bottom-up approach). Our framework is scalable in LLM selections and can be readily adapted to individual research needs \textbf{without} requiring the training of new models. The framework leverages the heterogeneity of diverse, open-source, and \textbf{locally-deployable LLMs} to enhance the accuracy and reliability of determining the presence or absence of a topic in a given text. The ensembling seeks a balance between the agreement and disagreement across multiple LLMs, guided by a relevancy scoring methodology. We evaluated the ensembling framework using both a public dataset related to a mental health condition and a protected clinical dataset, both complemented by human annotations. For the public dataset, we also employed the large-sized GPT-4o as a judge \cite{zheng_judging_2023}.  This framework was not designed to replace human evaluation but rather to alleviate the labor burden by providing human raters with a preliminary label and relevancy score for each text. Finally, this framework is uniquely accessible to researchers with fewer computational resources, as the LLMs being assembled can be run on consumer grade computers.

Another recent work utilized the ensemble of LLMs for data annotation tasks without training models, but our ensembling framework differs significantly from theirs. \citet{farr_llm_2024} proposed an ensembling methodology that aligns multiple LLMs in a chain, routing subsets of data to subsequent models based on the classification uncertainty measured for the current model. This ensemble is an extension of the LLM routing technique but does not require to train a router model. It was shown to outperform individual models in the chain and reduce the use of high-cost LLMs. Our ensembling framework differs from theirs in the following ways. First, as an extension of the LLM routing technique, their designs incorporated both open-source smaller-sized LLMs and a large-sized, closed-source GPT-4o. The extent to which performance improvements attributed to the use of GPT-4o remains unclear. Given the constraints of free-text data, our method mainly focuses on enhancing the performance of open-source deployable LLMs through ensembling. Second, their method suggests that more robust models be placed later in the chain, with the robustness indicated by a progression from smaller to larger model sizes. In our study, we aim to evaluate the advancements of ensembling compared to individual LLMs, controlling for the same model sizes. In addition to the F1-score assessed in their experiment, we evaluated the performance through multiple lenses, including precision, sensitivity and inter-rater reliability. Lastly, the labeling tasks evaluated in their empirical experiments—stance detection, misinformation detection, and ideology detection—are abstract components of text, that do not directly align with the needs for precise labeling of curated topics in psychological studies.


\begin{figure}
    \centering
    \includegraphics[width=1\linewidth]{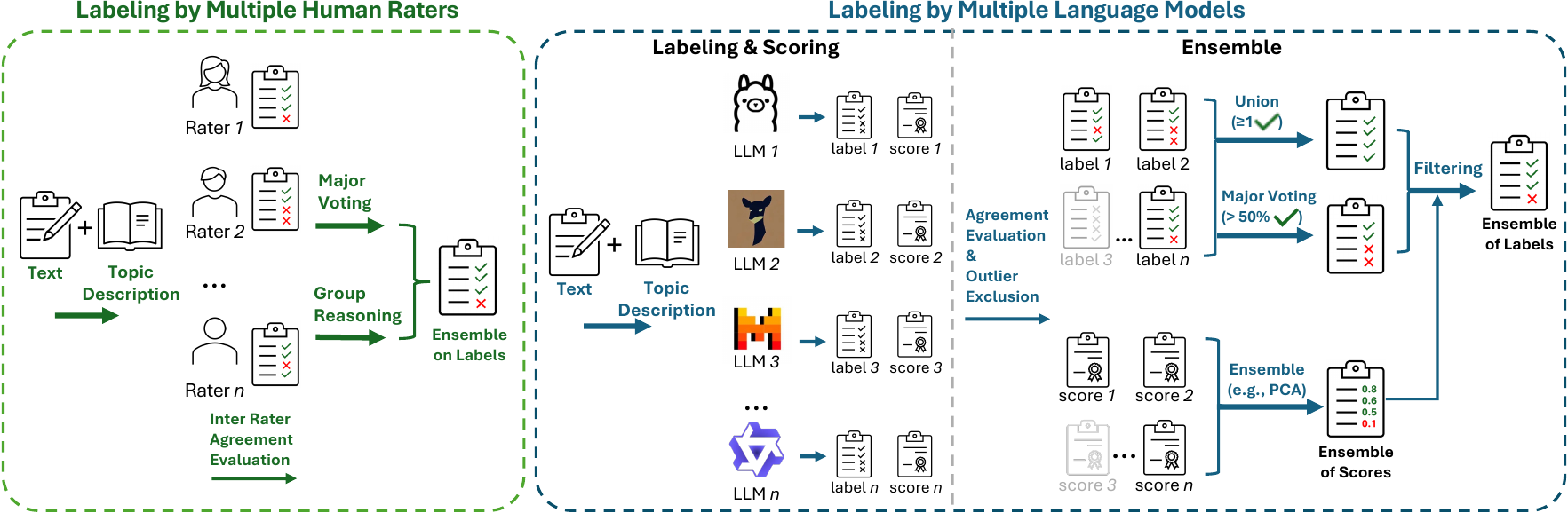}
    \caption{\small The ensemble framework of multiple LLMs (right) is analogous to the labeling process of multiple human raters (left) in the task of rating curated topics in free-text data. It leverages the heterogeneous performance of diverse open-source, deployable language models.
    (1) Labeling and scoring: Each model produces a label and a score for a given topic in an input text. Specifically, each model returns a dichotomous label indicating the presence of the topic, along with text phrases related to the topic that reflect its reasoning. A relevance score is then computed to measure the semantic similarity between the identified phrases and the predefined topic description. This is analogous to a human rater deciding whether a topic is present in a text (e.g., a patient narrative) and providing the phrases they identified as indicators.
    (2) Ensemble across LLMs: For a given topic, an ensemble score is calculated by averaging model scores (e.g., via PCA). The ensemble label is then determined by removing false positives from the union of positive cases across LLMs (excluding outliers), guided by the ensemble score. This mirrors majority voting combined with group reasoning among human raters to reach a final consensus. }
    \label{fig:framework}
\end{figure}

\section{Methods}\label{sec_method}


\subsection{Ensemble Framework} \label{sec_method_ensemble}

As shown in Figure \ref{fig:framework}, the ensemble framework of multiple LLMs is analogous to employing multiple human raters for rating curated topics in free-text data. In this section, we detail the key steps of the ensemble framework: (1) topic labeling by each selected open-source LLM; (2) relevance scoring based on the embedding distance between each LLM’s reasoning and the input topic description; (3) agreement evaluation across LLMs with the removal of potential outliers; and (4) ensembling of scores and labels from multiple LLMs.

\subsubsection{Topic Labeling}

A collection of open-source language models should be used to conduct the same labeling and scoring task for a topic of interest. In this experimentation, we selected four small-sized (7-8 billion parameters) language models that are open-source on huggingface: \textit{Llama-3.1-8B-Instruct} \cite{patterson_carbon_2022}, \textit{Qwen2.5-7B-Instruct} \cite{hui_qwen25-coder_2024}, \textit{Mistral-7B-Instruct-v0.3} \cite{jiang_mistral_2023}, and \textit{Vicuna-7b-v1.5} \cite{zheng_judging_2023, kassem_alpaca_2024}. Each model can be deployed in a secure computing environment and is recognized for its capacity in text generation inference \cite{li_llm_2024}.
The largest selected model requires a minimum of 20 GB full-precision VRAM, and our experiments were conducted on a single NVIDIA V100 GPU.

\begin{figure}
    \centering
    \includegraphics[width=1\linewidth]{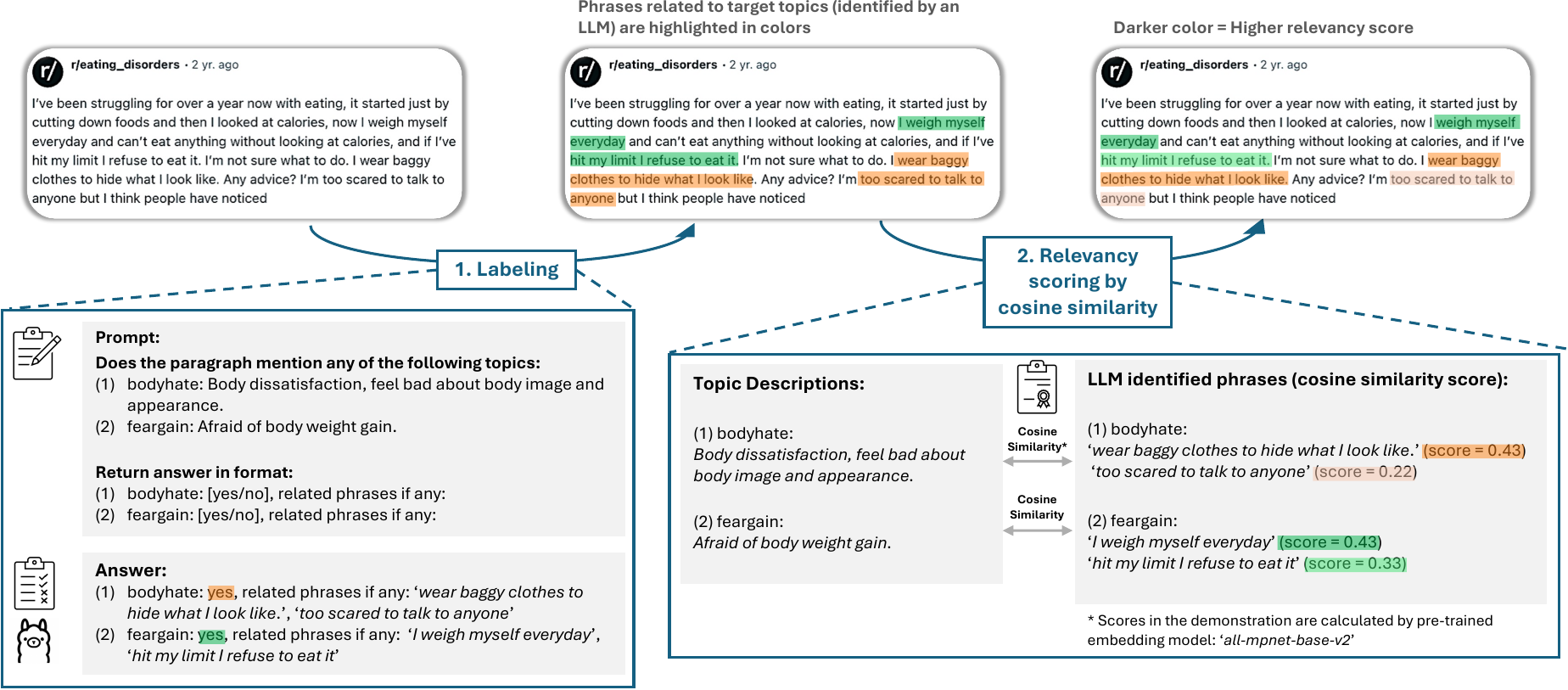}
    \caption{The procedure of topic labeling and relevance scoring performed by a single LLM, applied to an example text post from an eating disorders Reddit forum. Each LLM is prompted with the text, the target topics, and the corresponding detailed description of each topic. It then returns a dichotomous label indicating the presence of each topic, along with the text phrases identified as related to that topic, reflecting its reasoning. Then, a relevance score is computed for each topic by calculating the cosine similarity between the embedding of the topic description and the embedding of the returned phrases, thereby measuring the semantic relevance of that topic within the text. }
    \label{fig:label_and_score}
\end{figure}

To determine the presence of target topics in a given text, we prompted each LLM with the text and detailed descriptions of the topics, using the following template:
\begin{quote}
\texttt{\small
Does the paragraph mention any of the following topics: \\
(1) \{topic short name 1\}: \{description\}. \\
(2) \{topic short name 2\}: \{description\}.
... \\
Return answer in format: \\
(1) \{topic short name 1\}: [yes/no], related phrases if any: \\
(2) \{topic short name 2\}: [yes/no], related phrases if any: 
... \\
Paragraph: `\{text\}`
}
\end{quote}
Figure \ref{fig:label_and_score} illustrates an example of prompting a single LLM to determine whether the topics “body dissatisfaction (bodyhate)” and “afraid of body weight gain (feargain)” are present in a paragraph-length post from an eating disorder–related Reddit forum. The response from \textit{Llama-3.1-8B-Instruct} consists of a positive label (“yes”) along with related phrases for each topic—e.g., “wear baggy clothes to hide what I look like” and “too scared to talk to anyone” were identified as relevant to body image dissatisfaction. Although techniques such as few-shot prompting \cite{reynolds_prompt_2021}, chain-of-thought \cite{wei_chain--thought_2022}, and other prompt engineering techniques \cite{lan_stance_2024, ma_chain_2024} can be employed, this study focuses on comparing the ensemble performance with individual performances of LLMs in topic labeling tasks. We applied the same prompt template across all LLMs, allowing the models to reveal their heterogeneity in performing the same task.

\subsubsection{Relevancy Scoring}

Recent representation-learning–focused language models, exemplified by BERT (Bidirectional Encoder Representations from Transformers) and its variants \cite{devlin_bert_2019, liu2019roberta, reimers_sentence-bert_2019}, mark a significant advancement in capturing contextualized word, phrase, and sentence representations. These models generate numerical embedding vectors that encode the meaning of words in a given text, their contextual relationships, and the overall semantic context. The semantic distance between two texts (sentences or phrases) can be quantified using the cosine similarity between their vectors in the embedding space \cite{rahutomo2012semantic}.

Since each LLM is prompted to return topic-related phrases identified within the input text, we calculate a relevancy score for each topic by computing the cosine similarity between the embedding of the topic description and the embedding of the phrases returned by the LLM.
In the example shown in Figure~\ref{fig:label_and_score}, for the topic body dissatisfaction, the topic description provided in the prompt is “body dissatisfaction, feel bad about body image and appearance.” The cosine similarities between this description and the LLM-returned phrases are 0.43 for “wear baggy clothes to hide what I look like” and 0.22 for “too scared to talk to anyone.” By taking the maximum, the final relevancy score of topic bodyhate in this post is 0.43 for this LLM.
This score measures the extent to which the evidence provided by an LLM is relevant to the queried topic and reflects how explicitly the topic is discussed in the input text. 

Note that cosine similarity between two texts can differ across embedding models potentially due to the anisotropy of the embedding space \cite{li-etal-2020-sentence, su2021whitening}; some models produce similarities averaging around 0.5, while some yield averages closer to 0. For the generalizability of our framework across different embedding models, we calculate a baseline cosine similarity score between each topic description and an empty string, aiming to recalibrate similarity so that the distance to an empty string is centered at zero. Specifically, we adjust the similarity scores between the topic description and the returned phrases by subtracting this baseline and clamping negative values to zero, yielding a final relevancy score in the range [0, 1].

In our experiments, embeddings are generated using the pre-trained Sentence-BERT model \textit{all-mpnet-base-v2} \cite{reimers_sentence-bert_2019}, which maps long text or sentences (truncated to a maximum of 384 words) into a 768-dimensional vector space. This model, trained on over 1 billion sentences, has demonstrated optimal performance in semantic search according to the massive text embedding benchmark leaderboard \cite{muennighoff_mteb_2023}.

\subsubsection{Agreement Evaluation}

Before ensembling, we conducted an agreement evaluation on the labels and scores returned by the LLMs. Gwet's AC1 \cite{gwet_computing_2008} is a preferred inter-rater reliability metric in scenarios where the topic of interest is rarely prevalent \cite{chew_llm-assisted_2023}. Other metrics, such as Cohen's Kappa \cite{hsu_interrater_2003} and Fleiss's Kappa \cite{falotico_fleiss_2015}, can result in the ``high agreement, low reliability" paradox in rare-event labeling when the topic of interest is present at low prevalence \cite{feinstein_high_1990, falotico_fleiss_2015, zhao_assumptions_2013}. To evaluate the agreement among the continuous relevancy scores across LLMs, we categorized the scores into ten ordinal levels, each spanning 0.1, before calculating unweighted Gwet's AC1. We also reported Fleiss’s Kappa, as it is a commonly used inter-rater reliability coefficient for more than two raters (or LLMs in our case). The two coefficients are expected to differ because Gwet’s AC1 employs an adjusted chance agreement calculation to account for imbalances in category prevalence. Further details on Gwet’s AC1 and Fleiss's Kappa are provided in Appendix~\ref{ira}.

Although the ensemble of LLMs should benefit from the heterogeneity of diverse LLMs, those whose labels and scores significantly deviate from the rest may detrimentally affect the ensembling process. Thus, to remove potential outlier LLMs, we calculated the increase in Gwet's AC1, which is less sensitive to low topic prevalence, when each LLM was excluded from the full collection. If this increase exceeds a predetermined threshold (in our experiments, greater than 10\% of the overall AC1), the LLM is excluded from the ensembling candidates.

\subsubsection{Ensembling}

From each LLM, we retrieve a binary label and a relevancy score ranging from 0 to 1 if the label is positive. As shown in Figure~\ref{fig:framework}, we propose a simple ensembling procedure that consists of two stages: (1) ensemble of relevancy scores, and (2) ensemble of dichotomous labels. 

\begin{figure}
    \centering
    \includegraphics[width=\linewidth]{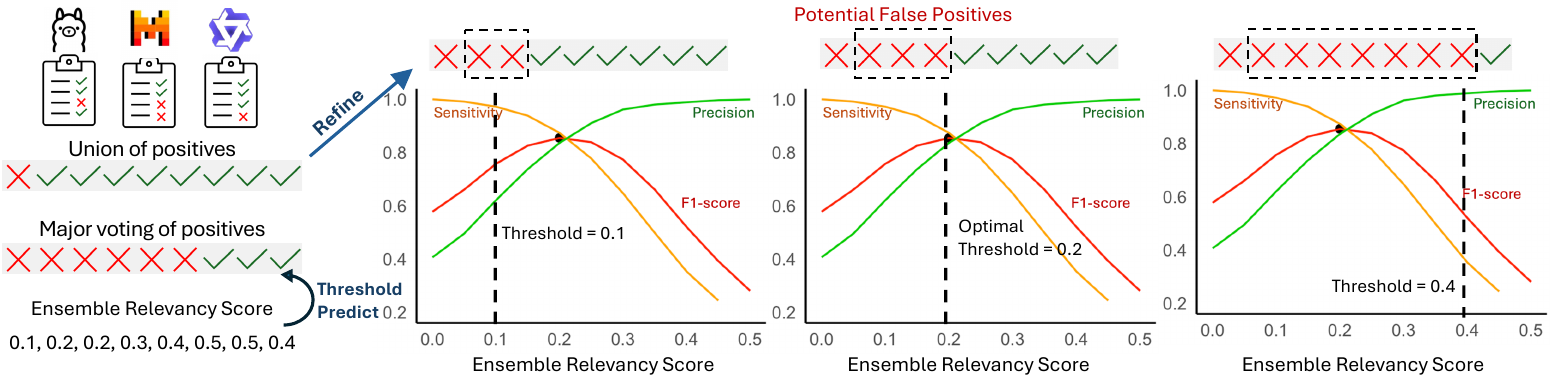}
    \caption{Process of the ensemble of labels by multiple LLMs. 
    Relevancy scores from multiple LLMs are aggregated using PCA to form a single ensemble score. 
    By sweeping thresholds on this scale, sensitivity, precision, and F1-score are computed against the intersection label (which is determined by the majority voting). 
    The optimal threshold maximizes F1 and positive texts with relevancy under the threshold are considered potential false positives and are reassigned as negatives in the final label.}
    
    \label{fig:ensemble_workflow}
\end{figure}

The ensemble of the relevancy scores aims to calculate a single relevancy of a topic in a text, aggregated across non-outlier LLMs. Compare to an unweighted average of scores across multiple LLMs, principal component analysis (PCA) looks for the linear combination of variables that explains the most variance in the data. 
For example, the principal component of four score variables (vectors of $N$ texts) $x_i$ from four LLMs is given by $\text{PC}_1 = w_1 x_1 + w_2 x_2 + w_3 x_3 + w_4 x_4,$ where the weights $w_i$ are chosen to maximize the variance of $\text{PC}_1$ subject to the normalization constraint $\sum_{i=1}^{4} w_i^2 = 1$. By achieving maximum variability in $\text{PC}_1$, we calibrate texts with mostly higher relevancy scores from LLMs to the upper end of the single ensemble score scale, and texts with mostly lower scores to the lower end.

For the ensemble of labels, we detailed the process in Figure~\ref{fig:ensemble_workflow}. 
The primary objective in ensembling the labels is to \textbf{eliminate potential false positives from the union of positive labels} returned by the LLMs. 
The union of positive labels from multiple LLMs inherits all the false positives from each LLM. Conversely, the intersection of positive labels---defined as those agreed upon by more than half of the LLMs, following the majority-voting convention commonly used in human rating processes---may serve as a closer proxy to the true label, but can potentially fail to include all true positives.
We term \textit{the union label} as the resulting label after taking the union of positive labels from multiple LLMs, and \textit{the intersection label} as the resulting label after taking the intersection (majority voting) of positives from multiple LLMs.
The ensembling of labels involves finding a balance between the union and the intersection (by majority voting), guided by the PCA ensemble of relevancy scores.
Specifically, we examined each threshold on the scale of the PCA ensemble of scores and identified the optimal threshold that \textit{predicts intersection labels} with the highest F1-score. 
For example, in Figure~\ref{fig:ensemble_workflow}, at each threshold on the scale of the PCA ensemble score, we compute a predicted label by classifying texts below the threshold as negative and those above as positive. We then calculate sensitivity and precision between the predicted label and the intersection label, defined as
\[
\text{Sensitivity} = \frac{\text{TP}}{\text{TP} + \text{FN}}, \quad
\text{Precision} = \frac{\text{TP}}{\text{TP} + \text{FP}},
\]
where TP, FP, and FN denote true positives, false positives, and false negatives, respectively. 
As the threshold moves from low to high on the PCA ensemble score scale, recall decreases while precision increases. 
We identify an optimal threshold where the F1-score (the harmonic mean of precision and recall),
\[
F1 = \frac{2 \cdot \text{Precision} \cdot \text{Recall}}{\text{Precision} + \text{Recall}},
\]
is maximized. 
This threshold represents the optimized trade-off between sensitivity and precision, particularly for topics with rare prevalence~\cite{hancock_evaluating_2023}.
Lastly, we modify the union label into a final ensemble label as follows: 
texts in the union of positive cases whose scores fall below the optimal threshold are considered potential false positives and are reassigned as negatives in the final label.

\subsection{Empirical Evaluation} \label{sec_method_evaluation}

Eating disorders (ED) are serious psychiatric illnesses that result in over 3.3 million healthy life years lost globally each year \cite{van_hoeken_review_2020}. In the area of LLM-aided mental health applications, the most frequently explored conditions include depression, suicide, and stress \cite{hua_large_2024}. Eating disorders are understudied, though a recent study \cite{chopra_deciphering_2024} that employed a closed-source LLM, `\textit{gpt-3.5-turbo-1106}', to analyze the psychosocial effects of ED on members of a Reddit forum. Yet, the study lacked evaluation by human or the use of multiple LLMs \cite{zheng_judging_2023}.

In this study, we evaluated the ensembling framework through two case studies of psychological textual data focused on eating disorders: (1) a public dataset of Reddit posts discussing topics related to dieting and ED, and (2) a private (protected) dataset describing ED patients’ experiences of weight stigma during treatment.
We used human annotation to evaluate the labeling performances of the four open-source LLMs previously listed, as well as the ensembles of all possible combinations of them. For the public dataset of Reddit posts, we also used closed-source large-sized GPT-4o as a judge \cite{zheng_judging_2023} to evaluate their labeling performances. The Reddit posts, along with a human-annotated subset, are made public through the University of Virginia Dataverse \cite{V3/NENELT_2025}. Note that human annotation is only used in evaluating the labels produced by individual LLMs or their ensembles; the ensemble framework does \textit{not} rely on human evaluation.

The study of weight stigma was approved by the Institutional Review Board (IRB) of University of Louisville (Protocol 20.0837). The study of Reddit posts used publicly available, de-identified data and was approved by the Institutional Review Board for the Social and Behavioral Sciences (IRB-SBS) of University of Virginia (Protocol 7572).

\begin{table*}
\caption{Descriptions of topic of interests in two empirical case studies}\label{table_topic_description}
\label{agreement_evaluation}
\subcaptionbox{\textbf{Case study 1}: labeling overlapping topics between dieting and ED discussions among reddit posts}{
\resizebox{0.9\textwidth}{!}{
\begin{tabular}[t]{ll>{\raggedright\arraybackslash}p{3.5in}}
\toprule
Topic & Short Name & Description\\
\midrule
Binge eating & binge & Binge eating.\\
Body dissatisfaction & bodyhate & Body dissatisfaction, feel bad about body image and appearance.\\
Calorie count & calorie & Count calorie.\\
Food cravings & crave & Craving for high calorie food or carbs.\\
Depressed mood & depressedmood & Depressed mood, feeling depressed.\\
\addlinespace
ED recovery & ed & Eating disorders(ED) diagnosis or recovery, ED includes anorexia nervosa, anorexic, bulimia, bulimic, binge eating disorders, arfid, osfed, pica.\\
Physical exercise & exercise & Physical exercise.\\
Fear foods & fearfood & Fear certain foods.\\
Fear of weight gain & feargain & Fear of body weight gain. Must related to body weight.\\
Weight gain & gain & Body weight gain.\\
\addlinespace
Weight loss & loss & Body weight loss.\\
High protein diet & protein & High protein diet, carbohydrate-reduced(low-carb) high-protein diet.\\
Relationships & relation & Family and social relationships.\\
Restriction & restrict & Restrict nutrition or calorie intake.\\
Ideal body image & idealbody & Ideal body image including thinness, skinny body, low body fat and lean body mass.\\
\bottomrule
\end{tabular}
}
}
\hfill

\subcaptionbox{\textbf{Case study 2}: labeling weight stigma experienced by ed patients during treatment}{
\resizebox{0.85\textwidth}{!}{
\begin{tabular}[t]{>{\raggedright\arraybackslash}p{5.5in}}
\toprule
Subcategories of Weight Stigma in Treatment\\
\midrule
Assumptions of health based on appearance or weight.\\
Healthcare decisions based on weight. Treatment decision based on weight.\\
Not sick enough, or not thin enough, or not low weight enough.\\
Not taken seriously because of weight, Inadequate care based on weight.\\
Diet promotion.\\
\addlinespace
Encouragement of weight loss.\\
Reassurance of thinness. Reassurance that will not get fat in treatment.\\
Negative attitudes, discrimination, or prejudice based on body weight or size.\\
Weight blamed for health issues or concerns.\\
Weight tied to personality characteristics.\\
\bottomrule
\end{tabular}
}
}
\hfill
\\
\end{table*}

\subsubsection{Case Study 1: Overlapping Topics Between ED and Diet Culture on Reddit}

Social media has shown to promote diet culture, characterized by health myths about food and eating, as well as a moral hierarchy of bodies \cite{jovanovski_demystifying_2022, harrison2018diet}. Though diet culture lacks a uniform definition \cite{jovanovski_demystifying_2022}, it is defined in this study as a system of beliefs that equates thinness with health and oppresses people who do not align with standards of body shape, size, and weight \cite{harrison2018diet}. Research shows that dieting and other weight-control methods significantly increase the risk of harmful eating practices associated with disordered eating or ED \cite{hilbert_risk_2014}. In this case study, we aimed to identify potentially overlapping topics between ED and general dieting online discussions to better understand the similarities between diet culture and ED. Extended from previous analyses of ED online discussions \cite{moessner_analyzing_2018, punzi_network-based_2022}, we identified 15 candidate topics that are likely prevalent in both dieting and ED communities, as listed in Table \ref{table_topic_description} (a).

\begin{wraptable}{r}{0.6\textwidth} 
\centering
\caption{Case study 1 -- Data Description}\label{case1_data_description}
\centering
\resizebox{\linewidth}{!}{
\begin{tabular}[t]{lll>{\raggedright\arraybackslash}p{2.5in}}
\toprule
 & Forums & Posts (Authors) & Search by\\
\midrule
ED & 12 & 14853 (8878) & eat disorder, anorexia nervosa, bulimia nervosa, binge eating disorder, ARFID, avoidant/restrictive food intake disorder, pica, other specified feeding or eating disorders, OSFED, unspecified feeding or eating disorders, orthorexia\\
Dietary & 56 & 34367 (25459) & diet, food, nutrition, meal, eat, fat, calorie, ingredient, menu\\
Fitness & 47 & 27955 (20200) & body shape, body image, body weight, thinness, exercise, fit, obesity\\
\bottomrule
\end{tabular}}
\end{wraptable}

We collected textual posts from 115 Reddit forums across three topical groups—dietary, fitness, and eating disorders—using the Python package PRAW (v7.7.1). For each forum, we retrieved 1000 top-rated posts and 1000 newest posts, retaining unique entries. Posts in dietary and fitness forums contain content of general dieting online discussions. In total, we collected 77,175 posts from 53,784 authors across 115 subreddit forums. Data descriptions and the designated searching keywords are provided in Table \ref{case1_data_description}. For ED forums, we included 12 subreddits: \textit{`EatingDisorders', `bulimia', `intuitiveeating', `AnorexiaNervosa', `EDanonymemes', `EDAnonymous', `BingeEatingDisorder', `fuckeatingdisorders', `ARFID', `eating\_disorders', `save\_food', `edsupport'}. The data acquisition took place in February 2024.

As a public dataset, we conducted the evaluation in this case study using both human annotators and the large-size, closed-source GPT-4o. Firstly, for practicality, we randomly sampled 1,080 posts—10 posts per forum from 108 forums that contains at least 10 posts. Two authors independently labeled all topics across the 1,080 posts. The Gwet’s AC1 and Fleiss’ Kappa scores on the 1,080 posts labeled by the two coders were 0.972 (95\% CI: 0.966 - 0.979) and 0.851 (95\% CI: 0.817 - 0.886), respectively, indicating excellent agreement and reliability of the human annotation. Secondly, we used GPT-4o to label and rate all 77,175 posts. The Gwet’s AC1 and Fleiss’ Kappa with human annotation were 0.976 (95\% CI: 0.974 - 0.979) and 0.885 (95\% CI: 0.873 - 0.897), respectively, suggesting that large-sized LLMs can achieve close-to-human performance in these tasks, as previously recognized \cite{khraisha_can_2023, gargari_enhancing_2024}.

\subsubsection{Case Study 2: ED Patients' Experiences with Weight Stigma}

In the second case study, we used a \textbf{protected} clinical dataset of ED patients narratives about their experience with weight stigma during treatment. A total of 1,368 texts from patients treated for eating disorders were collected within a clinical setting. 
Two domain experts independently coded the dataset, determining the presence of weight stigma in patient narratives by identifying any of the subcategories listed in Table \ref{table_topic_description} (b). The Gwet’s AC1 and Fleiss’ Kappa between two coders' annotation were 0.625 (95\% CI: 0.583 - 0.667) and 0.541 (95\% CI: 0.494 - 0.588), respectively. Due to the relatively low human inter-rater reliability, we used the subset of 1,080 texts where both coders agreed as the final human annotation to evaluate the labeling performance of LLMs and their ensembles.

In this case study, the LLMs' labeling and scoring process resembled that of human labeling and was based on a codebook of weight stigma \cite{xiao_supporting_2023}. Each LLM was prompted to determine the presence of each subcategory, using the same detailed descriptions provided to human coders in the codebook. Subsequently, the label of weight stigma was calculated based on the presence of any subcategory, while the relevancy score was determined by the average relevancy scores of the present subcategories.

\section{Results}\label{sec_result}

\begin{table*}[b]
\caption{Evaluation of agreement on labeling and scoring\textsuperscript{1} conducted with four open-source, small-sized language models\textsuperscript{2}. Bold values indicate higher agreement in scores than in labels returned by language models.}
\label{agreement_evaluation}
\subcaptionbox{\textbf{Data 1}: ED and Dieting Reddit Posts}{
\resizebox{\textwidth}{!}{
\begin{tabular}{>{\raggedright\arraybackslash}p{0.9in}>{\raggedright\arraybackslash}p{1.4in}>{\raggedright\arraybackslash}p{1.4in}>{\raggedright\arraybackslash}p{1.4in}>{\raggedright\arraybackslash}p{1.4in}}
\toprule
\multicolumn{1}{c}{ } & \multicolumn{2}{c}{Gwet's AC1 [95\% CI]} & \multicolumn{2}{c}{Fleiss' kappa [95\% CI]} \\
\cmidrule(l{3pt}r{3pt}){2-3} \cmidrule(l{3pt}r{3pt}){4-5}
Topic & Labels & Scores & Labels & Scores\\
\midrule
binge & 0.939 [0.937, 0.940] & 0.924 [0.922, 0.925] & 0.651 [0.644, 0.659] & \textbf{0.702} [0.695, 0.709]\\
bodyhate & 0.827 [0.825, 0.830] & \textbf{0.852} [0.850, 0.854] & 0.451 [0.445, 0.457] & \textbf{0.484} [0.477, 0.492]\\
calorie & 0.900 [0.898, 0.902] & 0.881 [0.879, 0.883] & 0.671 [0.665, 0.676] & \textbf{0.722} [0.718, 0.727]\\
crave & 0.858 [0.856, 0.860] & \textbf{0.884} [0.882, 0.886] & 0.354 [0.347, 0.361] & \textbf{0.408} [0.398, 0.418]\\
depressedmood & 0.774 [0.771, 0.777] & \textbf{0.854} [0.852, 0.856] & 0.321 [0.316, 0.325] & \textbf{0.442} [0.435, 0.450]\\
\addlinespace
ed & 0.902 [0.900, 0.904] & 0.879 [0.877, 0.881] & 0.689 [0.683, 0.694] & \textbf{0.758} [0.753, 0.763]\\
exercise & 0.832 [0.829, 0.835] & 0.781 [0.779, 0.784] & 0.730 [0.726, 0.734] & 0.647 [0.643, 0.651]\\
fearfood & 0.946 [0.944, 0.947] & 0.946 [0.944, 0.947] & 0.300 [0.291, 0.309] & \textbf{0.354} [0.340, 0.369]\\
feargain & 0.904 [0.902, 0.905] & 0.903 [0.901, 0.905] & 0.377 [0.369, 0.385] & 0.360 [0.351, 0.369]\\
gain & 0.880 [0.878, 0.882] & 0.864 [0.862, 0.866] & 0.563 [0.557, 0.568] & \textbf{0.573} [0.567, 0.579]\\
\addlinespace
idealbody & 0.790 [0.787, 0.792] & \textbf{0.828} [0.825, 0.830] & 0.246 [0.240, 0.251] & 0.232 [0.225, 0.238]\\
loss & 0.751 [0.748, 0.754] & \textbf{0.772} [0.769, 0.774] & 0.546 [0.542, 0.551] & \textbf{0.596} [0.591, 0.601]\\
protein & 0.796 [0.793, 0.799] & \textbf{0.820} [0.818, 0.823] & 0.452 [0.447, 0.458] & \textbf{0.596} [0.589, 0.603]\\
relation & 0.338 [0.333, 0.342] & \textbf{0.825} [0.823, 0.827] & 0.057 [0.053, 0.061] & \textbf{0.350} [0.343, 0.357]\\
restrict & 0.761 [0.757, 0.764] & \textbf{0.793} [0.791, 0.796] & 0.495 [0.490, 0.500] & \textbf{0.526} [0.521, 0.532]\\
\bottomrule
\end{tabular}
}
}
\hfill
\subcaptionbox{\textbf{Data 2 (Protected)}: ED Patients Experiences}{
\resizebox{\textwidth}{!}{
\begin{tabular}{>{\raggedright\arraybackslash}p{0.9in}>{\raggedright\arraybackslash}p{1.4in}>{\raggedright\arraybackslash}p{1.4in}>{\raggedright\arraybackslash}p{1.4in}>{\raggedright\arraybackslash}p{1.4in}}
\toprule
\multicolumn{1}{c}{ } & \multicolumn{2}{c}{Gwet's AC1 [95\% CI]} & \multicolumn{2}{c}{Fleiss' kappa [95\% CI]} \\
\cmidrule(l{3pt}r{3pt}){2-3} \cmidrule(l{3pt}r{3pt}){4-5}
Topic & Labels & Scores & Labels & Scores\\
\midrule
weightstigma & 0.601 [0.572, 0.629] & 0.597 [0.575, 0.618] & 0.187 [0.152, 0.222] & \textbf{0.300} [0.269, 0.330]\\
\bottomrule
\end{tabular}
}
}
\hfill
\\
\begin{minipage}{1\textwidth}
\footnotesize{$^1$ Embedding scores were calculated using \textit{all-mpnet-base-v2} and binned into 10 ordinal levels for agreement evaluation.}

\footnotesize{$^2$ The four open-
source, small-sized language models are \textit{Llama-3.1-8B-Instruct}, \textit{Qwen2.5-7B-Instruct}, \textit{Mistral-7B-Instruct-v0.3}, and \textit{Vicuna-7b-v1.5}.}
\end{minipage}
\end{table*}

\subsection{Mitigating Heterogeneity through Relevancy Scoring}

As shown in Table \ref{agreement_evaluation}, in both case studies, the relevancy scores from LLMs demonstrated higher agreement for most topics, as measured by Gwet's AC1 and Fleiss' kappa, than the dichotomous labels returned by the LLMs. This result suggests that (1) the scoring method can effectively mitigate the heterogeneity in the labeling decisions made by multiple LLMs, and (2) the reasoning of LLMs exhibited higher similarity than their binary classifications.  Gwet’s AC1 and Fleiss’s Kappa differ because they calculate chance agreement differently: Gwet’s AC1 uses an adjusted chance agreement to account for imbalances in category prevalence, whereas Fleiss’s Kappa relies directly on marginal distributions.

\subsection{Enhancing Accuracy through Ensembling}

\begin{figure}
    \centering
    \includegraphics[width=1\linewidth]{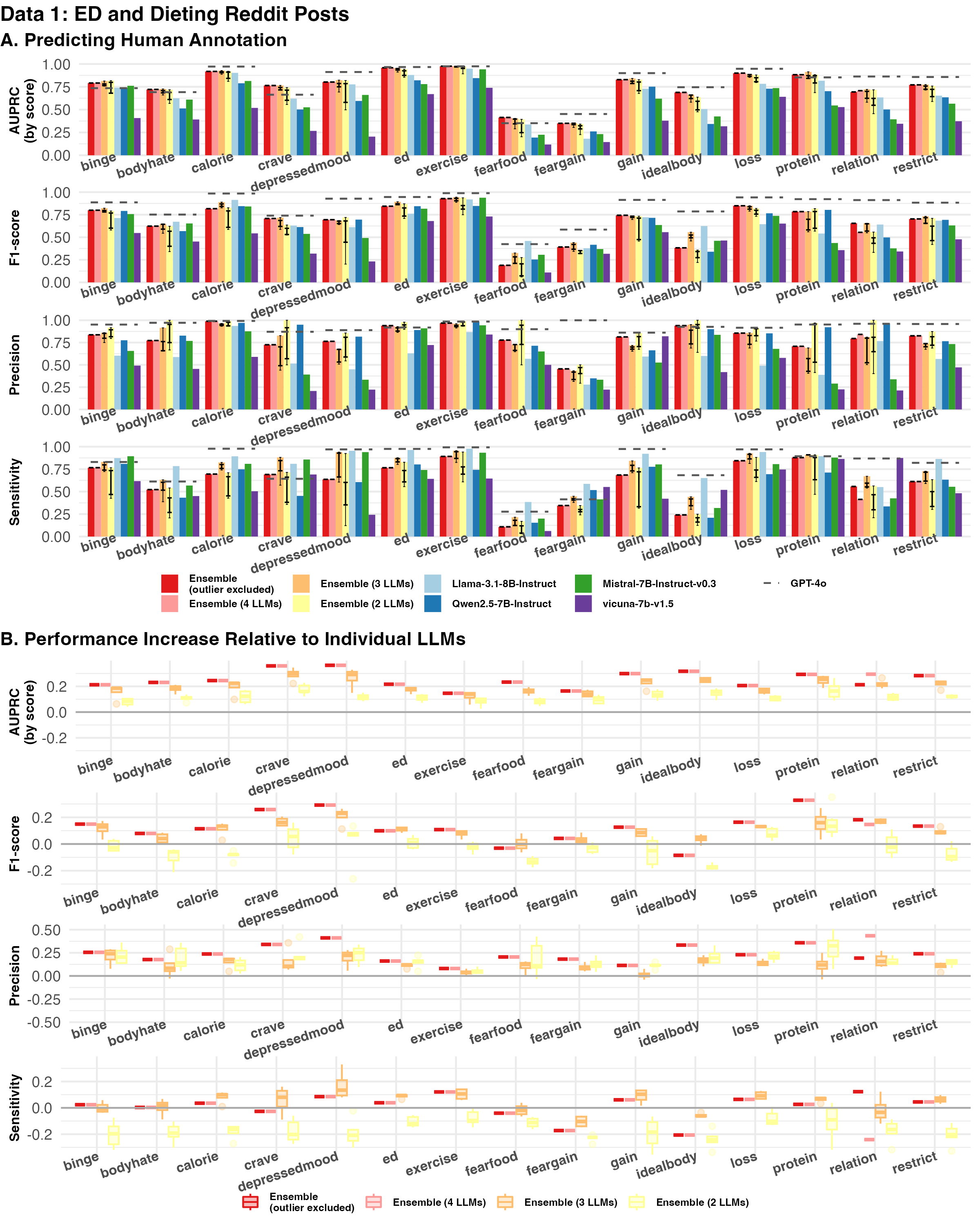}
    \caption{Performance in predicting human labels of 15 topics related to eating disorders and dieting in Reddit posts, using four open-source, locally-deployable LLMs and all combinations of their ensembles. \textbf{Panel A.} illustrates the performance of using relevancy scores or labels returned by LLMs and their ensembles to classify human annotations. The AUPRC metric evaluates relevancy scores, while F1-score, precision, and sensitivity evaluate performance based on labels. Different colors represent various LLMs and their ensembles. Performance metrics for GPT-4o are indicated by horizontal dashed lines. For ensembles of two and three LLMs, error bars indicate the minimum, maximum, and the 25th and 75th percentiles across all combinations of the respective ensemble size. \textbf{Panel B.} illustrates the performance increases achieved by using ensembles of LLMs to classify human annotations, compared to the median performance of the individual LLMs included in each ensemble.}
    \label{fig:human_evaluation1}
\end{figure}

\textbf{In case study 1}, as shown in Figure \ref{fig:human_evaluation1}, when using same-sized open-source LLMs to label overlapping topics between ED and dieting online discussions, there was large heterogeneity in their performances. Specifically, Qwen2.5-7B-Instruct exhibited low sensitivity but high precision, indicating it tended to be too conservative in predicting positive cases, yielding more false negatives. Conversely, Llama-3.1-8B-Instruct displayed high sensitivity but low precision, suggesting it was too generous in predicting positive cases, resulting in more false positives. Mistral-7B-Instruct-v0.3 had varying performances across different topics, while Vicuna-7b-v1.5 showed the lowest performance across all metrics.

However, the ensembles of LLMs overcame the shortcomings of individual LLMs and exhibited the highest performance in predicting human annotations across almost all topics. Firstly, the ensemble of four LLMs' labels exhibited higher F1-score, sensitivity and precision than individual LLMs. Similarly, the PCA ensembles of scores showed the highest Area Under the Precision-Recall Curve (AUPRC) compared to individual LLMs. Specifically, the ensemble of four LLMs and the ensemble with outliers removed (final ensemble) showed significantly higher precision than all LLMs except for Qwen2.5-7B-Instruct (who had the lowest sensitivity), suggesting that ensembling can effectively identify and remove false positives. On the other hand, the 4-LLMs ensemble and the final ensemble exhibited higher sensitivity than all LLMs except for Llama-3.1-8B-Instruct (who had the lowest precision), suggesting that ensembling can effectively incorporate more true positives according to the agreement by multiple LLMs. Secondly, as shown in panel B, ensembling a larger number of LLMs resulted in higher performance gains, while using only two LLMs in the ensemble tended to be vulnerable to the heterogeneous performances of individual LLMs. Lastly, the accuracy of GPT-4o was generally higher than that of the selected open-source LLMs, as indicated by the dashed horizontal lines in Panel A. Using all 77,175 posts, the ensemble of LLMs demonstrated performance closer to GPT-4o compared to individual LLMs, as depicted in Figure \ref{fig:gpt4o_evaluation1}. This result also suggests that the parameter size of an LLM should be considered an important factor in evaluating LLM-assisted tasks. 

\begin{figure}
    \centering
    \includegraphics[width=1\linewidth]{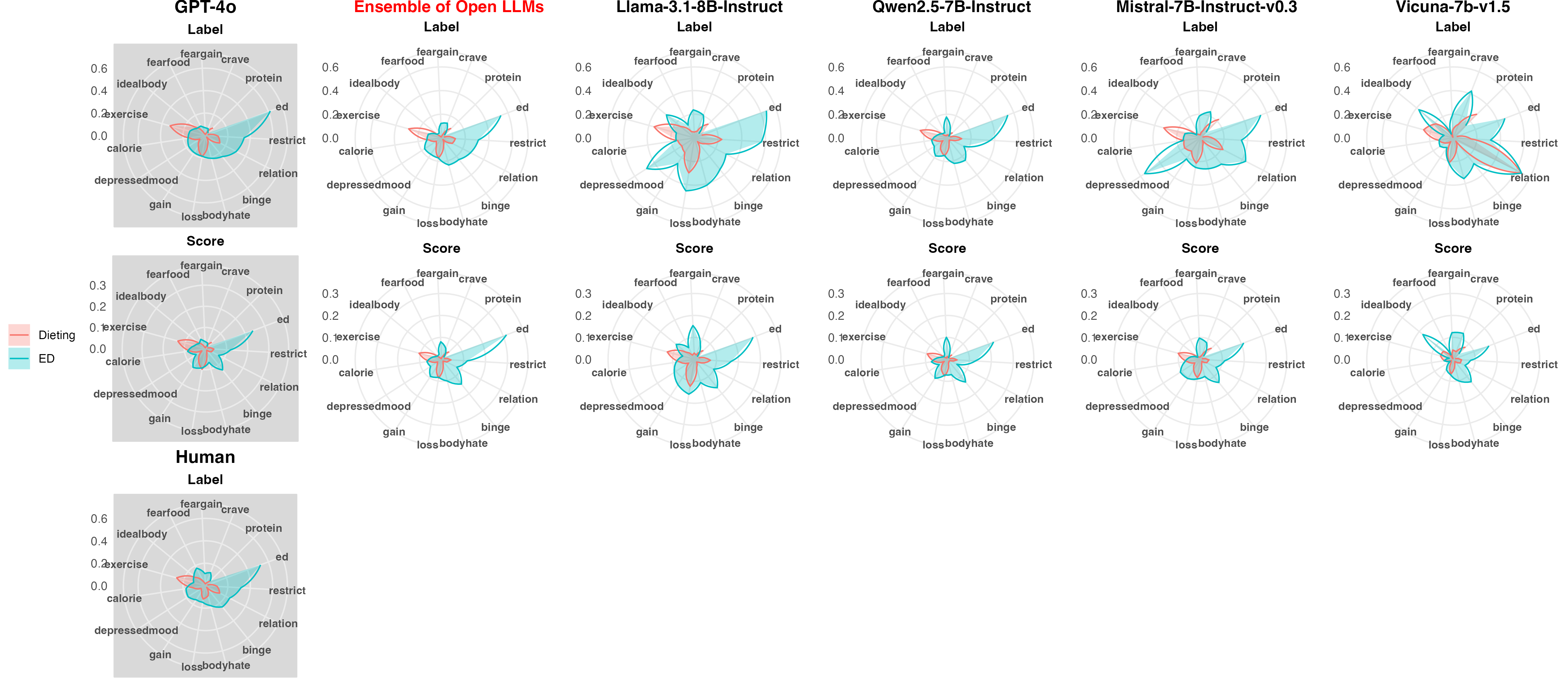}
    \caption{The average occurrence rates and average relevancy scores of 15 ED and dieting topics, as returned by LLMs and the ensemble, stratified by two forum groups: ED forums (blue) and dieting (dietary or fitness) forums (red). For comparison, the average occurrence rates labeled by GPT-4o and human annotators are highlighted with a gray background. }
    \label{fig:spider1}
\end{figure}

Figure \ref{fig:spider1} illustrates another comparison between the ensemble and individual LLMs, based on the distribution of their labels and scores across 77,175 posts. These are aggregated by the topical group of forums, with 'ED' indicating eating disorder forums and 'dieting' referring to dietary and fitness forums. Notably, the distribution of the average occurrence rate and relevancy score by the ensemble of LLMs most closely resembled those by GPT-4o and humans, compared to labels and scores by individual LLMs. There were large discrepancies in the distribution of labels and scores returned by individual LLMs.

\begin{figure}
    \centering
    \includegraphics[width=1\linewidth]{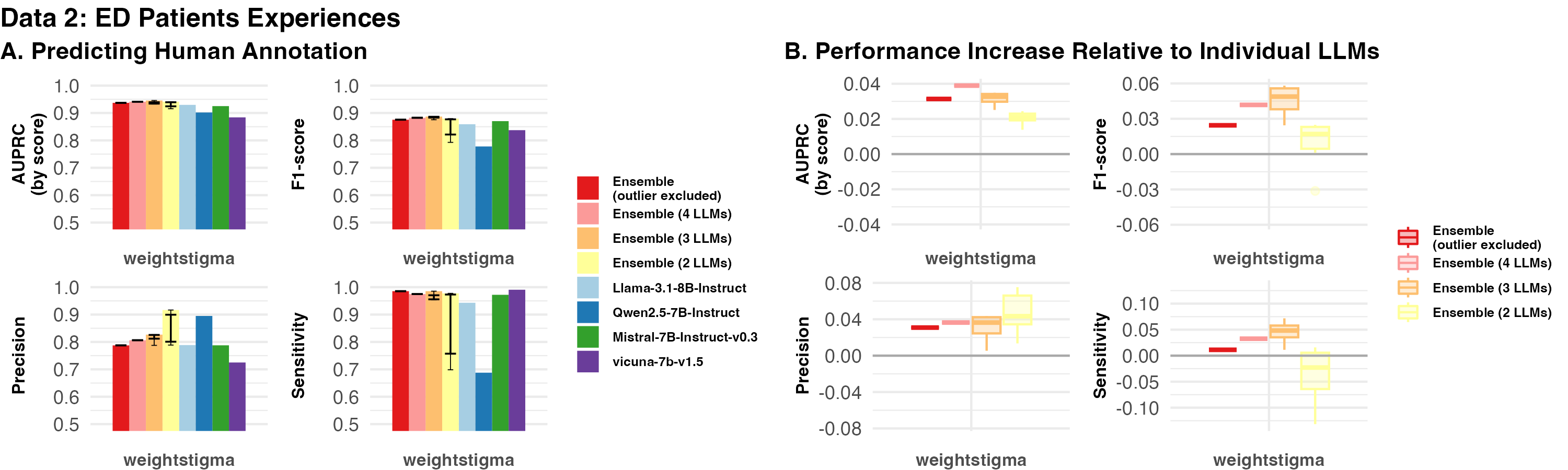}
    \caption{Performance in predicting human annotations of weight stigma experienced by ED patients during treatment, using four open-source, locally-deployable LLMs and all combinations of their ensembles. \textbf{Panel A.} illustrates the performance of using relevancy scores or labels returned by LLMs and their ensembles to classify human annotations. The AUPRC metric evaluates relevancy scores, while F1-score, precision, and sensitivity evaluate performance based on labels. Different colors represent various LLMs and their ensembles. For ensembles of two and three LLMs, error bars indicate the minimum, maximum, and the 25th and 75th percentiles across all combinations of the respective ensemble size. \textbf{Panel B.} illustrates the performance increases achieved by using ensembles of LLMs to classify human annotations, compared to the median performance of the individual LLMs included in each ensemble.}
    \label{fig:human_evaluation2}
\end{figure}

\textbf{In case study 2}, as shown in Figure \ref{fig:human_evaluation2}, LLMs exhibited varying capabilities in determining the presence of weight stigma in ED patients' narratives. Qwen2.5-7B-Instruct was too conservative in labeling positive cases, resulting in low sensitivity and high precision, whereas Vicuna-7b-v1.5 was too generous, resulting in high sensitivity but low precision. Nevertheless, the ensemble of three or more LLMs achieved higher F1-score, precision, and sensitivity than individual LLMs when classifying human annotations. Similarly, the ensembles of scores showed the highest AUPRC compared to individual LLMs. The removal of the detected outlier LLM, Qwen2.5-7B-Instruct, led to increased sensitivity but slightly decreased precision. The overall F1-score remained the same as that of the ensemble of four LLMs. 

\section{Discussion} \label{sec_discussion}

In psychology research, annotating granular topics, such as pathological behaviors and mentalities in free-form texts like patient narratives, is typically labor-intensive and time-consuming. Without user consent for external use, LLM-assisted labeling techniques that rely on closed-source models through APIs can hardly be applied to such free-text data. These constraints also limit the use of large-sized LLMs ($>$100 billion parameters) because they cannot be deployed locally in most research settings.

In this study, we proposed an ensembling framework that leverages the heterogeneity of diverse, open-source, locally-deployable LLMs for topic labeling and relevancy scoring on free-text data. This ensembling is guided by a relevancy scoring methodology that utilizes embedding distances between the topic description and LLMs' reasoning. The topic labeling and scoring by multiple LLMs mirror the human labeling process performed by several human coders. The main idea of ensembling labels from multiple LLMs is to eliminate potential false positives from the union of positive labels returned by multiple LLMs, which inherits all false positives from each LLM. Conversely, the intersection of positive labels, based on majority voting among LLMs, may be closer to the ground truth but is likely to exclude true positives. 
Thus, our label ensembling seeks a balance between the union and the intersection, guided by the ensemble of relevancy scores.
The framework was not designed to replace human evaluation, but rather to alleviate the burden of the human labeling process by providing a preliminary label and a relevancy score for each text. The final labeling decision, typically made by domain experts, can reference these preliminary labels, focusing particularly on positive labels with low relevancy scores (potential false positives) and negative labels with positive scores (potential false negatives).

In the experimentation, we deployed four small-sized (7-8 billion parameters) LLMs locally: \textit{Llama-3.1-8B-Instruct, Qwen2.5-7B-Instruct, Mistral-7B-Instruct-v0.3, Vicuna-7b-v1.5,} and evaluated the framework through two case studies on human-annotated psychological textual data, with a focus on eating disorders (ED). The first study analyzed overlapping topics between ED and dieting across 77,175 public Reddit posts, while the second documented ED patients' experiences with weight stigma during treatment in a free-text dataset of 1,080 patient narratives. We found that: 

\begin{enumerate}
    \item Evaluated by human annotations, there is heterogeneity among the selected same-sized LLMs when conducting the same topic labeling tasks, with some exhibiting low sensitivity but high precision, and others high sensitivity but low precision. Such heterogeneity reflects the varied data, model architectures, and preprocessing steps each LLM was exposed to during training, as well as constraints imposed by the size of their parameters. 
    
    \item The ensemble of LLMs effectively leveraged the heterogeneity of the LLMs, achieving the highest accuracy in predicting human annotations. In the first case study, the ensemble of LLMs also demonstrated the highest accuracy in predicting closed-source large-sized GPT-4o's labels. The improvement in accuracy was evidenced by superior F1-score, superior precision, and sensitivity compared to individual LLMs. Moreover, the ensemble of a larger number of LLMs demonstrated more reliable performance improvements and a more optimal precision-recall trade-off. In contrast, ensembles of only two LLMs appeared more vulnerable to the variability in individual LLM performance.

    \item The relevancy scores across LLMs showed greater agreement than dichotomous labels, as measured by Gwet's AC1 and Fleiss' kappa. This suggests the proposed relevancy scoring method can effectively mitigate the variability in LLMs' labeling.
    
\end{enumerate}

We also observed heterogeneity in the performance of LLMs when annotating and rating different topics. Although the ensembles generally outperform individual LLMs, for some topics, all LLMs exhibited worse performance, as measured by F1, precision, and recall. This might be due to the complexity of the topic definitions. For example, topics such as 'fear of certain foods' and 'fear of weight gain' posed challenges not only for open-source small-sized LLMs but also for GPT-4o to detect. These topics required understanding both the emotional content, 'fear,' and the context of food or weight gain from the text. Similarly, the definition of 'ideal body image' is complex; it may involve being muscular and having lean body mass for males, while for females, it often pertains to being thin and skinny. Although the ensembles maintained higher performances than individual LLMs, enhancing LLMs' understanding of topic definitions through more complex prompting techniques, such as chain of thought prompting \cite{wei_chain--thought_2022} or soft prompting \cite{lester2021power}, or fine-tuning on datasets annotated with these topics, are important next steps.

Nevertheless, given the proposed ensembling framework, potential extensions at each stage warrant further investigation. First, while this study focused on ensembling open-source, deployable LLMs for free-text data, the framework could be expanded to include any textual data and larger-sized LLMs. For example, we controlled for the parameter size of LLMs to be between 7-8 billion parameters in our experiments. Larger-sized open-source LLMs, such as Llama-3.1-70B-Instruct and Llama-3.1-405B \cite{patterson_carbon_2022}, may warrant examination in future studies that have access to more computational resources. Second, techniques such as few-shot prompting \cite{reynolds_prompt_2021}, chain-of-thought \cite{wei_chain--thought_2022}, and other prompt engineering techniques \cite{lan_stance_2024, ma_chain_2024} can be employed as alternatives for the simple prompting in this experimentation. The potential of performance improvement by advanced prompting may result in improved accuracy for both individual LLMs and the ensemble of them. Third, the relevancy scoring utilizes the embedding distance between topic description and LLM-returned related phrases. Besides the pre-trained Sentence-BERT model `all-mpnet-base-v2,' other pre-trained embedding models, whether encoder-decoder or decoder-only, may serve as potential alternatives. Lastly, an intriguing direction for future study is to fine-tune individual LLMs based on the ensemble of labels and scores for a given topic. This approach may provide individual models with unseen information and bridge their capacity to perform inference tasks in specific psychological contexts.




\clearpage
\begin{appendices}
\appendix

\section{Inter-rater agreement coefficients}\label{ira}

\subsection{Gwet's AC1}
Gwet's AC1 \cite{gwet_computing_2008} is an inter-rater agreement coefficient designed to overcome the limitations of Cohen's or Fleiss' Kappa, particularly their sensitivity to prevalence and marginal distribution. Like Kappa, AC1 is defined as the ratio of observed agreement beyond chance over the maximum possible agreement beyond chance:
\[
AC1 = \frac{P_o - P_e^*}{1 - P_e^*},
\]
where $P_o$ is the observed proportion of agreement among raters, and $P_e^*$ is the expected agreement by chance. 

Let $n_{ij}$ denote the number of raters who assigned item $i$ to category $j$, with $n$ raters, $N$ items and $k$ categories in total. The observed agreement $P_o$ is given by
\[
P_o = \frac{1}{N} \sum_{i=1}^N \sum_{j=1}^k \frac{n_{ij}(n_{ij} - 1)}{n(n-1)},
\]

Let
$p_j = \frac{1}{N} \sum_{i=1}^N \frac{n_{ij}}{n}$
be the overall proportion of ratings assigned to category $j$. The \textbf{adjusted chance agreement} $P_e^*$ is given by 
\[
P_e^* = \frac{1}{k-1} \sum_{j=1}^k p_j (1 - p_j).
\]
This adjustment prevents artificially low agreement coefficients when one category is dominant, making AC1 more robust in cases of imbalanced class distributions. For comparison, Fleiss's Kappa is defined in the next section.

\subsection{Fleiss's Kappa}

Fleiss's Kappa \cite{falotico_fleiss_2015, fleiss1981measurement} is a generalization of Cohen's Kappa for assessing inter-rater agreement among $n\ge 2$ raters when assigning categorical ratings to $N$ items. Like other chance-corrected coefficients, it is defined as
\[
\kappa = \frac{P_o - P_e}{1 - P_e},
\]
where $P_o$ is the observed agreement and $P_e$ is the expected agreement by chance.

Let $n_{ij}$ denote the number of raters who assigned item $i$ to category $j$, with $n$ raters, $N$ items and $k$ categories in total. 
The observed agreement $P_o $ across all items is given by
\[
P_o = \frac{1}{N} \sum_{i=1}^N  \sum_{j=1}^k \frac{n_{ij}(n_{ij} - 1)}{n(n-1)}.
\]

The \textbf{chance agreement} $P_e$ is given by 
\[
P_e = \sum_{j=1}^k \left( \frac{1}{N} \sum_{i=1}^N \frac{n_{ij}}{n} \right)^2.
\]
Although widely used, the chance agreement $P_e$ is sensitive to prevalence and marginal distributions, which can result in the ``high agreement, low reliability'' paradox when the target category has low prevalence.

\begin{figure}
    \centering
    \includegraphics[width=1\linewidth]{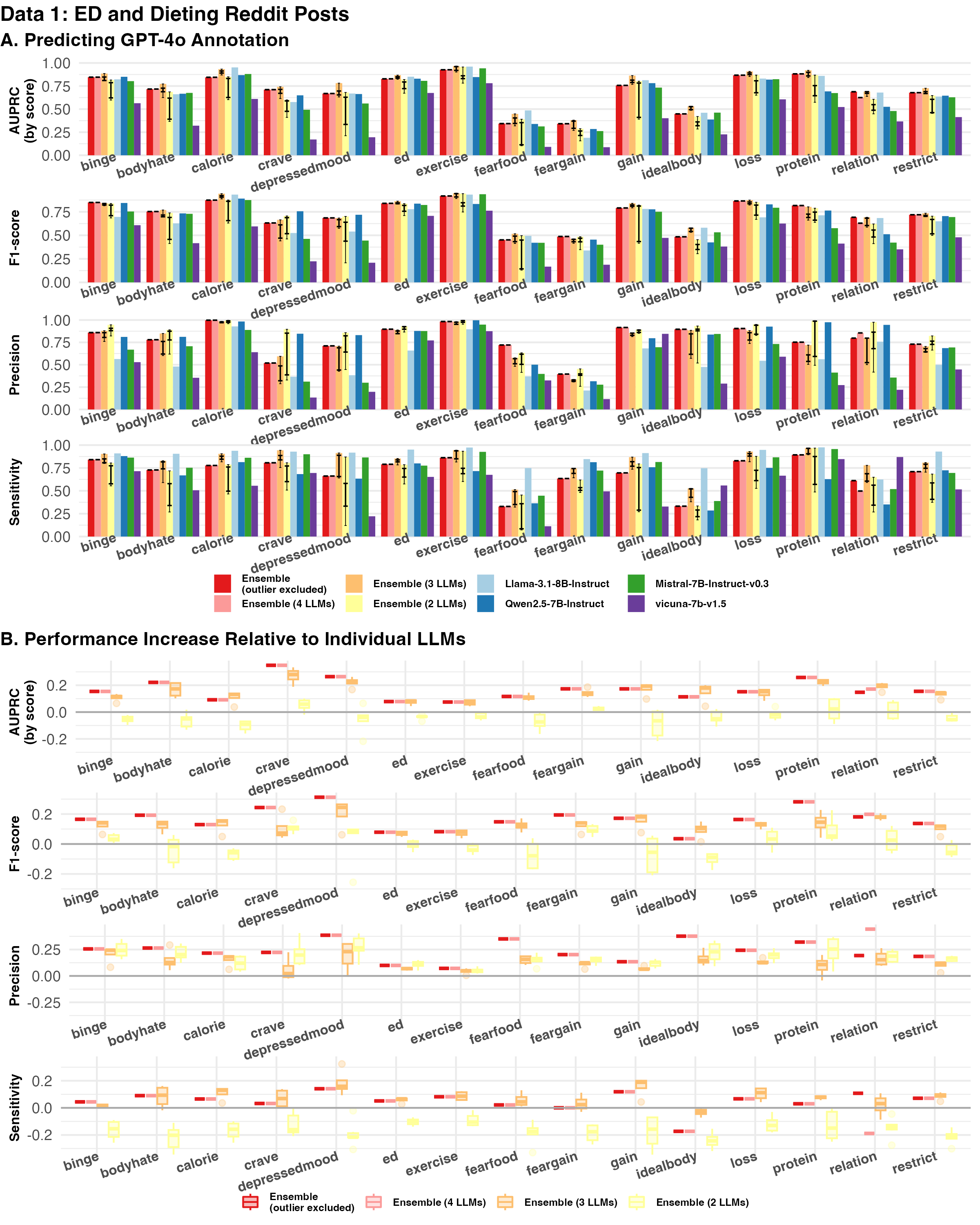}
    \caption{Performance in predicting GPT-4o labels of 15 topics related to eating disorders and dieting in Reddit posts, using four open-source, locally-deployable LLMs and all combinations of their ensembles. \textbf{Panel A} illustrates the performance of using relevancy scores or labels returned by LLMs and their ensembles to classify GPT-4o labels. The AUPRC metric evaluates relevancy scores, while F1-score, precision, and sensitivity evaluate performance based on labels. Different colors represent various LLMs and their ensembles. For ensembles of two and three LLMs, error bars indicate the minimum, maximum, and the 25th and 75th percentiles across all combinations of the respective ensemble size. \textbf{Panel B} illustrates the performance increases achieved by using ensembles of LLMs to classify GPT-4o annotations, compared to the median performance of the individual LLMs included in each ensemble.}
    \label{fig:gpt4o_evaluation1}
\end{figure}

\end{appendices}

\clearpage
\bibliography{ref}

\end{document}